\documentclass[accept,copyright]{eptcs}
\providecommand{\event}{CREST 2019} 
\usepackage{breakurl}             
\usepackage[english]{babel}
\usepackage[utf8]{inputenc}
\usepackage{graphicx}
\usepackage{subcaption}
\usepackage{amsmath}
\usepackage{dashbox}
\usepackage{amsmath}
\usepackage{framed}
\usepackage{hyperref}
\usepackage{todonotes}
\usepackage{listings}

\newtheorem{definition}{Definition}

\usepackage{wrapfig}

\title{Extending Causal Models from Machines into Humans}
\author{Severin Kacianka\qquad Amjad Ibrahim\qquad Alexander Pretschner
\institute{Technical University of Munich\\ Munich, Germany}
\email{firstname.lastname@tum.de}
\and
Alexander Trende \qquad Andreas L\"{u}dtke 
\institute{Offis\\
Oldenburg, Germany}
\email{firstname.lastname@offis.de}
}
\def\titlerunning{Extending Causal Models from Machines into Humans}
\def\authorrunning{S. Kacianka, A. Ibrahim, A. Pretschner, A. Trende \&
A. L\"{u}dtke}
\begin{document}
\maketitle

\begin{abstract}

Causal Models are increasingly suggested as a means to reason about the behavior
of cyber-physical systems in socio-technical contexts. They allow us to analyze
courses of events and reason about possible alternatives. 
Until now, however, such reasoning is confined to the technical domain and
limited to single systems or at most groups of systems. The humans that are an
integral part of any such socio-technical system are usually ignored or dealt
with by ``expert judgment''. 
We show how a technical causal model can be extended with models of human
behavior to cover the complexity and interplay between humans and technical
systems.  This integrated socio-technical causal model can then be used to
reason not only about actions and decisions taken by the machine, but also about
those taken by humans interacting with the system. 
In this paper we demonstrate the feasibility of merging causal models about
machines with causal models about humans and illustrate the usefulness of this
approach with a highly automated vehicle example.  

\end{abstract}

\section{Introduction}

\begin{figure}%
    \centering
    \begin{subfigure}{3cm}
        \includegraphics[height=4cm]{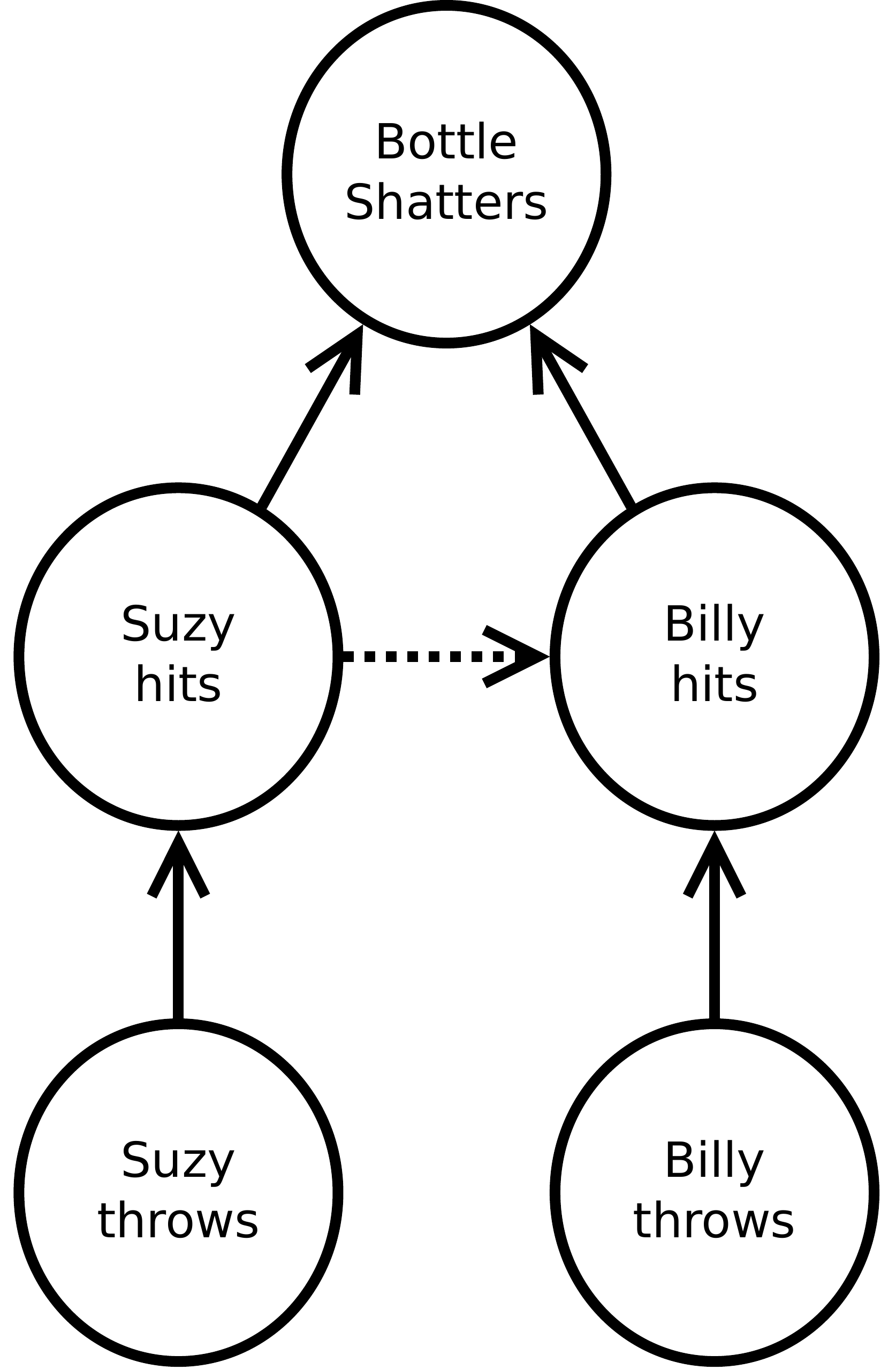}
        \caption{Classic version.}	\label{fig:rock}
    \end{subfigure}
    \qquad
    \begin{subfigure}{5cm}
        \includegraphics[height=4cm]{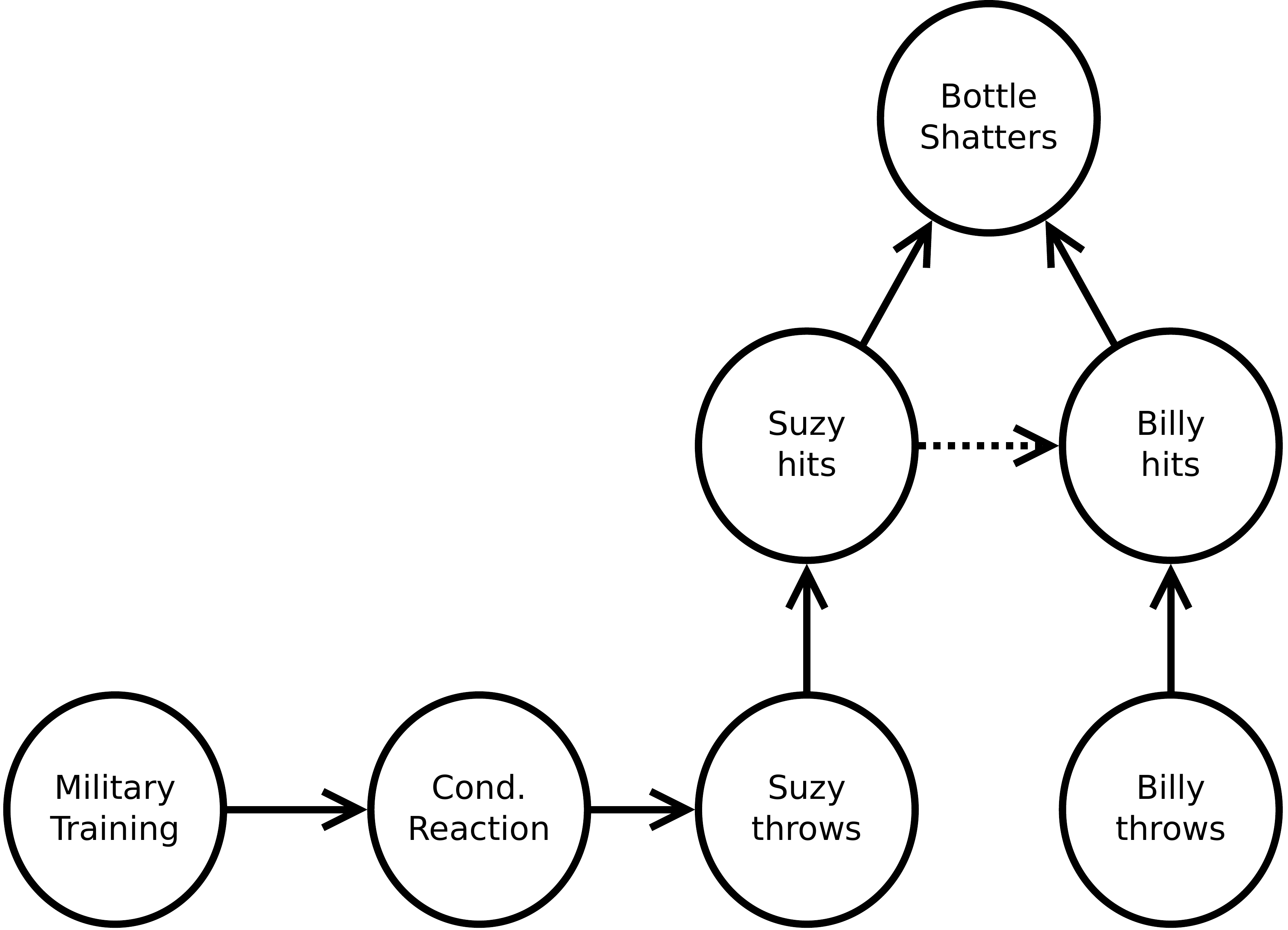}
        \caption{With military conditioning.}
        \label{fig:conditioning}
    \end{subfigure}
    \caption{The rock-throwing example.}%
\end{figure}

Causality is an essential building block to answer complex questions of
accountability \cite{kacianka2018understanding}. It allows us to reason about
different courses of actions and go beyond mere correlation based reasoning.
While building causal models is still a major problem and technologies to
automatically build causal models are still in their infancy, recent research
has shown that it is possible to derive useful starting points for models from
system descriptions \cite{nfm}. One can, for example, use fault trees as a
starting point for a causal model and then ask experts to refine it. While this
still requires human intervention, it is a lot easier to improve models than to
create them from scratch.  However, most Cyber-Physical Systems~(CPS) are
embedded into larger Socio-Technical Systems (STS) and interact with people. In
current research, automatically generated causal models only encompass the
technical aspects of these STS. Models for the interaction between humans and
machines are usually created by experts in an ad-hoc fashion. The field of human
factors is concerned exactly with this problem (e.g.,
\cite{weber2009modellierung}): how to model the behavior of a human that
interacts with a technical system (e.g., a pilot or a driver). The result of
this research are different models of human cognition and human behavior.  We
show how these models can be converted to causal models and linked with the
technical causal models. 
To illustrate the general idea, we use the classic rock throwing thought
experiment from the causality literature. In this scenario, two persons, Billy
and Suzy, will throw rocks at a bottle. If Suzy's rock hits the bottle slightly
before Billy, who is the cause for the bottle to shatter?  This example is
constructed  to show that simple counter factual reasoning is not enough to
attribute causality: Had Suzy not thrown the rock, the bottle would still have
shattered (because Billy also threw a rock), therefore Suzy cannot be the cause
for the bottle to shatter. Finding definitions of causality that will not lead
to such counterintuitive results culminated in the Halpern-Pearl definition of
causality \cite{halpern2005causes1}. It resolves the issues by introducing a so
called preemption relation \cite{nfm} that can express that Suzy's throw
preempted Billy's throw (see Figure~\ref{fig:rock}). 

While such a model expresses the objective facts very well, we cannot reason
about why Suzy threw faster. To do so, we would need a model of Suzy's mind. If,
for example, she was a soldier and, as part of her training, was conditioned to
throw rocks the moment a bottle appeared in her field of view\footnote{Soldiers
are conditioned to shoot at human shaped targets without second thought. This
increases their ``efficiency'' on the battlefield \cite{grossman2001killing}.},
we might not simply say that ``Suzy throwing the rock caused the bottle to
shatter'', but extend our causal chain to ``Her military training caused Suzy to
automatically throw the rock at the bottle, shattering it''. Instead of just
blaming Suzy, we could also consider her military training. 

Returning to CPS and their interaction with humans, we can now utilize existing
models of human behavior, as in Figure~\ref{fig:conditioning}, transform them to
causal models and link them with causal models of the technical systems. Instead
of just saying ``The car crashed, because the driver pressed the red button'',
we can say that ``Drivers are conditioned to press the red button in an
emergency; this lead to the driver pressing the button and the car crashing''.
While it is true that we might not have enough data in many cases, in the cases
where we do have enough data, like the example in Section~\ref{sec:example}, and
can actually extend the causal model into the human mind, we can gain valuable
insights and avoid the unsatisfying  generic answer ``human error''.  

In this paper we investigate the \emph{problem} of joining causal models of
technical systems with causal models of their operators and people they interact
with. As a \emph{solution} we propose to extend current methods to derive causal
models from system models to also include models of the human behavior. Our
\emph{contribution} is a detailed example of such an integrated socio-technical
causal model based on an automotive use case, preliminary research into a
process of converting human models to causal models and showing how these
integrated models can be used to reason over accidents.

\section{Background}
\subsection{The Preliminaries for Causal Models and Inference}
In this present paper, as in previous work \cite{nfm}, we build on the  Halpern
and Pearl (HP for short) definition of causality
\cite{halpern2015,halpern2005causes1,halpern2005causes2,pearl2018book}. This
definition models the world as two distinct sets, called  \textit{endogenous}
and \textit{exogenous}  variables. Their relation is described with a set of
\textit{structural equations}. Exogenous variables model factors outside of the
model and  endogenous variables model our understanding of the causal relations
within our model. The value of endogenous variables is
determined by the exogenous variables and the structural equations.
\cite{halpern2015} defines the conditions under which an endogenous variable can
be the cause of a specific state of another variable in the model.
Definition~\ref{def:cm} provides the formal definition of  binary causal models. 

\begin{definition} \label{def:cm}
	\textbf{Binary Causal Model} \cite{halpern2015}\\
	A binary causal model $M$ is a tuple $M = \mathcal{((U,V,R),F)}$, where
	\begin{itemize}
	    \item $\mathcal{U}$ is a set of exogenous variables,
		\item $\mathcal{V}$ is a set of endogenous variables, 
		\item $\mathcal{R}:$ associates with every variable  the nonempty set  
		$\{0,1\}$ of possible values,
		\item $\mathcal{F}$ associates with each variable $X \in
		\mathcal{V}$ a function that determines the value of $X$ (from the set 
		of possible values $\mathcal{R}(X)$) given the values of all other 
		variables\\
		$F_X : (\times_{U \in \mathcal{U}}\mathcal{R}(U)) \times (\times_{Y \in
			\mathcal{V}-\{X\}}\mathcal{R}(Y)) \to \mathcal{R}(X)$. 
	\end{itemize}

\end{definition}

A \textit{primitive event}, given $\mathcal{(U,V,R)}$, is a formula of the
form $X = x$ for $X \in \mathcal{V}$ and  $x \in \mathcal{R}(X)$. \textit{A
causal formula} is of the form $[Y_1 \leftarrow y_1, \dots, Y_k	\leftarrow
y_k]\varphi$, where $\varphi$ is a Boolean combination of primitive events.
$Y_1,\dots,Y_k$ (abbreviated $\overrightarrow{Y}$) are distinct variables in
$\mathcal{V}$, and $y_i \in \mathcal{R}(Y_i)$. Intuitively, this notation says
that  $\varphi$ would hold if $Y_i$ were set to $y_i$ for each $i$. $(M,
\overrightarrow{u}) \models X = x$ if the variable $X$ has value $x$ in the
unique solution to the equations in $M$ in context $\overrightarrow{u}$. An
intervention on a model is expressed either by setting the values of
$\overrightarrow{X}$ to $\overrightarrow{x}$, written as $[X_1 \leftarrow x_1,
.., X_k \leftarrow x_k]$, or by fixing the values of $\overrightarrow{X}$ in the
model, written as $M_{\overrightarrow{X} \leftarrow \overrightarrow{x}}$. So,
$(M, \overrightarrow{u}) \models [\overrightarrow{Y} \leftarrow
\overrightarrow{y}]\varphi$ is identical to $(M_{\overrightarrow{Y} \leftarrow
\overrightarrow{y}},\overrightarrow{u}) \models \varphi$.

\begin{definition}\label{def:ac}
	\textbf{Actual Cause} (latest/modified version
	\cite{halpern2015})\\ $\overrightarrow{X} = \overrightarrow{x}$
	is an actual cause of $\varphi$ in $(M,\overrightarrow{u})$ if the following
	three conditions hold:\\ \textbf{AC1.} \hspace{2mm} $(M,\overrightarrow{u})
	\models (\overrightarrow{X} = \overrightarrow{x})$ and
	$(M,\overrightarrow{u}) \models \varphi$.\\ \textbf{AC2.} \hspace{2mm} There
	is a set $\overrightarrow{W}$ of variables in $\mathcal{V}$ and a setting
	$\overrightarrow{x}'$ of the variables in $\overrightarrow{X}$ such that if
	$(M,\overrightarrow{u}) \models \overrightarrow{W} = \overrightarrow{w}$,
	then $(M,\overrightarrow{u}) \models [\overrightarrow{X} \leftarrow
	\overrightarrow{x}', \overrightarrow{W} \leftarrow \overrightarrow{w}] \neg
	\varphi$.\\ \textbf{AC3.} \hspace{2mm} $\overrightarrow{X}$ is minimal.
\end{definition}

\begin{figure}%
    \centering
    \begin{subfigure}{2.5cm}
        \includegraphics[height=2.5cm]{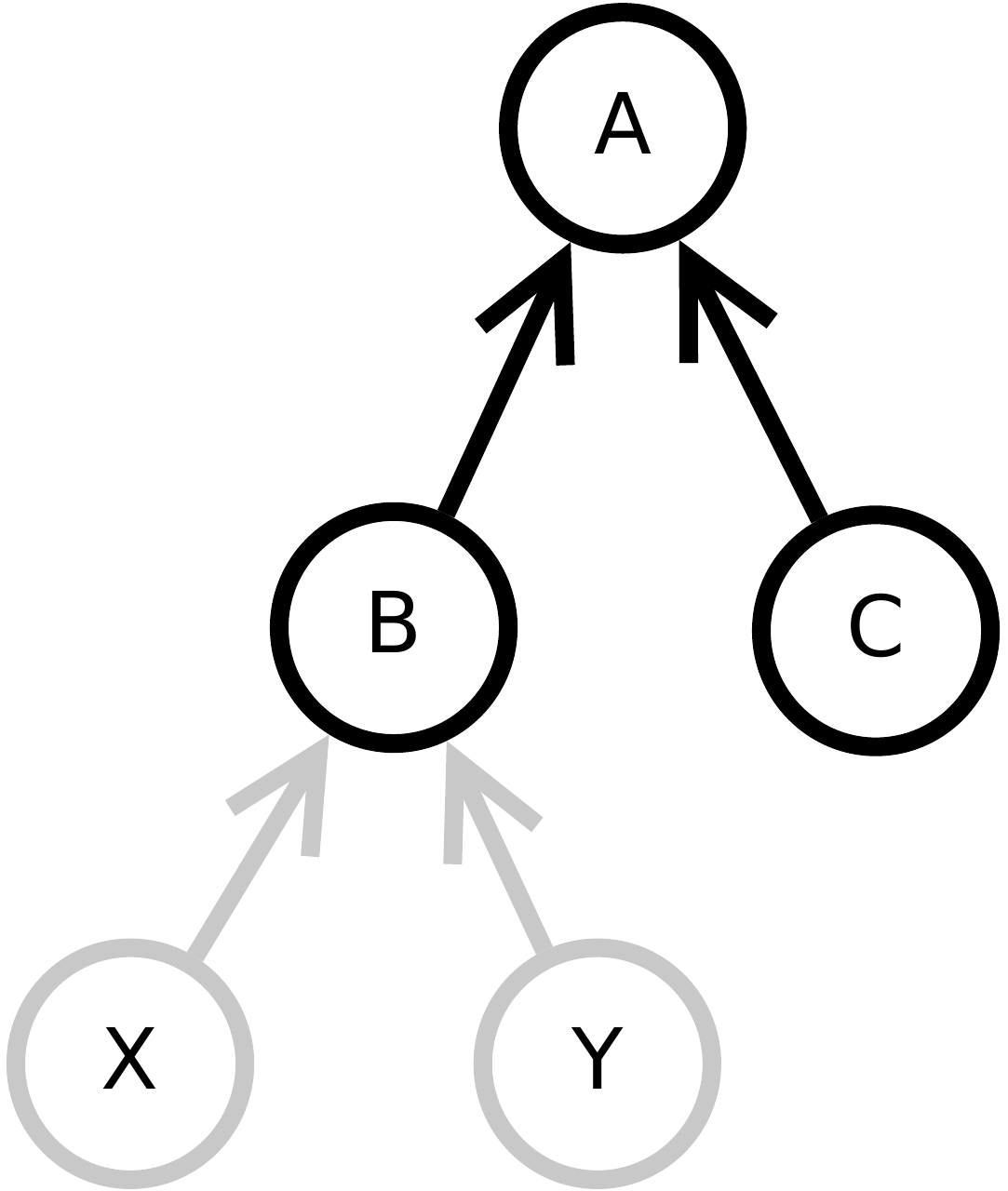}
    \subcaption{Refine.}
    \label{fig:refine}
    \end{subfigure}
    \qquad
    \begin{subfigure}{2.5cm}
        \includegraphics[height=2.5cm]{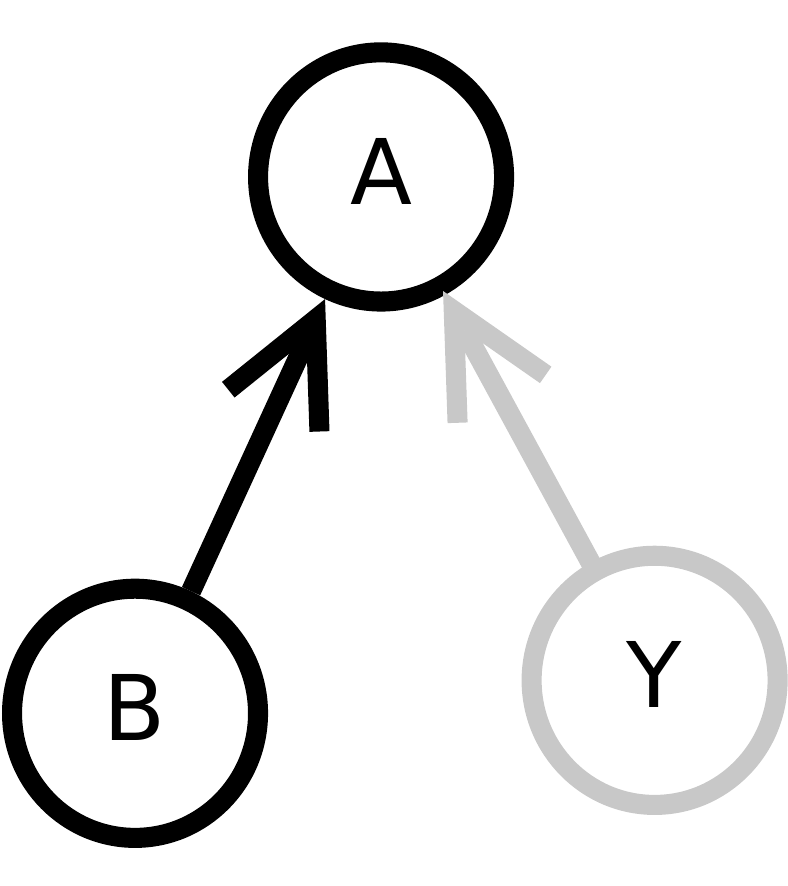}
		\subcaption{Extend.}
		\label{fig:extend}
    \end{subfigure}
     \caption{Merging Causal Models \cite{nfm}.}%
	\label{fig:merge_models}
\end{figure}
\subsection{Joining Causal Models}

For joining causal models, we rely on previous work \cite{nfm}. To briefly
summarize the method, it utilizes Alrajeh et al. \cite{alrajeh2018combining} and
Friedenberg and Halpern \cite{friedenberg2018combining} to automatically combine
models that are either  \textit{compatibility} or extended with a \emph{focus
function}. Without going into the technical details here (see \cite{nfm} for an
in-depth discussion), if one model explains more about cause and effect
relations than another model and both models talk about the same things, we can
simply combine those models. In the simplest case, one model \emph{refines} a
leaf node of an existing model (see Figure~\ref{fig:refine}).  In such a case, 
both models are compatible and can be automatically combined.
\emph{Extensions} (see Figure~\ref{fig:extend}) are unfortunately much more
complex. In some scenarios they can be joined automatically, but often they will
require the intervention from an expert. 

\subsection{Technical Source Models}
In this paper, we focus on fault and attack trees as technical source models.
Fault trees are used in the domain of safety, reliability, and risk assessment
methodology \cite{vesely1981fault,kordy2014dag} and are a means to analyze a
system's resilience against faults at design time \cite{ruijters2015fault}. They
are usually presented graphically as trees that model component faults that
might lead to system failure and their relationship among each other. While the
exact syntax varies, fault trees definitions usually include  \textit{AND},
\textit{OR}, \textit{EXCLUSIVE OR}, \textit{PRIORITY AND} and \textit{INHIBIT}
gates. 

Attack Trees stem from the field of security research
\cite{schneier1999attack,schneier2004secrets} and are used to model potential
security threats within a system and possible ways to exploit them. Similar to
fault trees, they are graphically represented as trees. The ultimate goal of an
attack is the root node of the tree and steps necessary to achieve that goal are
notated in sub-nodes. Attack trees usually support \textit{OR} and \textit{AND}
gates,

As in previous work \cite{nfm}, we follow Mauw and Oostdijk's
\cite{mauw2005foundations} formalization of attack trees, but replace their
multi-set semantics with Bruns and Anderson's propositional semantics for fault
trees \cite{bruns1993validating} and extend them with the semantics for fault
trees. This semantic matches the semantics of binary
structural equations used in causal models. Therefore, each non-leaf node in the
tree is expressed with a propositional formula of its parents, e.g., $out=in_1
\land in_2$. 

\begin{definition}\label{def:aft}
\textbf{Attack/Fault Tree} \\
	 $A(F)T$ is a 3-tuple $A(F)T = \mathcal{(N,\to, }n_0)$ where  $\mathcal{N}$
	 is a finite set of nodes,  $n_0 \in \mathcal{N}$ is the root node and
	 $\rightarrow \subseteq \mathcal{N \times N}$ is a finite set of 
	 acyclic relations. 
\end{definition}

\subsection{Human Behavior Source Models}

\begin{wrapfigure}{r}{.4\linewidth}
    \centering
 	\includegraphics[width=0.38\textwidth]{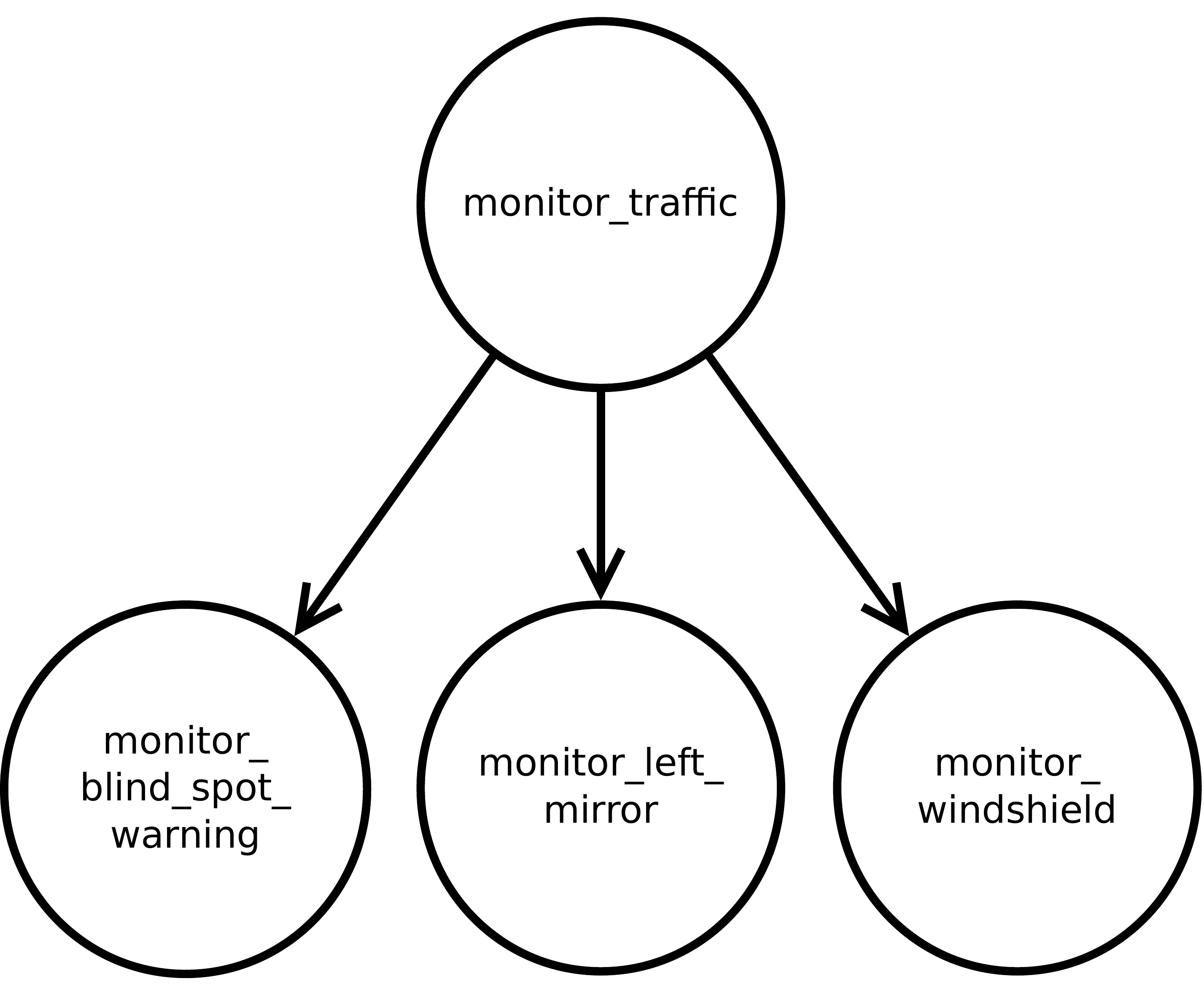}
	\caption{Sub-goal dependencies. }	
	\label{fig:goal_example}
\end{wrapfigure}
In the following, Hierarchical Task Analysis (HTA) \cite{annett2003hierarchical}
will be explained and how it can be used to create models of human behavior. The
HTA is used to partition tasks into sub-tasks and to demonstrate their
dependencies. The partition into sub-tasks is done in a way that finishing the
sub-tasks leads to completing the main task. Overall the abstract model consists
of the main task, sub-tasks, plans and operations and is expressed in a tree
diagram in most cases. An integral part of an HTA is the definition of the task
and the data collection. Different types of data collection are possible: e.g.,
questionnaires, experiments, or safety records. 

Models of human behavior can be constructed based on the results of an HTA. A
first step could be to construct a state chart. This reflects the order and
interplay of the previously defined tasks. Afterwards the HTA or state chart
could be translated into a model usable in a cognitive architecture.
\cite{kassner2011hierarchical} performed an HTA for lane merging maneuvers onto
freeways to define and improve the structure of the cognitive driver model
defined in \cite{weber2009modellierung}.  Cognitive architectures like ACT-R
\cite{anderson1997act} or CASCaS \cite{ludtke2010cognitive} use a rule-based
format for their model descriptions.  In the domain of cognitive architectures
the above described tasks and sub-tasks are called goals and sub-goals. This
terminology will be used in the following sections.  Sub-goals defined by an HTA
can be formulated as procedures in this format.  Every time step of the
simulation the cognitive cycle will be simulated. In this process a currently
active goal is being selected and executed. This will most likely trigger the
next sub-goal in the model. We will show the structure and procedure of such a
simulation while using four small sub-goals of an HTA: We assume that three
sub-goals were defined as part of a model describing a successful merging
maneuver: \emph{observe\_blind\_spot\_warning}, \emph{monitor\_traffic},
\emph{observe\_windshield} and \emph{observe\_left\_mirror}. The corresponding
dependencies between the sub-goals are shown in Figure~\ref{fig:goal_example}.

This means that the sub-goal \emph{monitor\_traffic} can be separated into the
sub-goals \emph{observe\_blind\_} \emph{spot\_warning},
\emph{observe\_windshield} and \emph{observe\_left\_mirror}.  Translated into
the rule-based format usable in CASCaS:

\begin{lstlisting}
rule(goal=monitor_traffic){ 
  Condition(boolean expression) 
 --> 
  Goal(observe_blind_spot_warning) 
  Goal(observe_windshield) 
  Goal(observe_left_mirror) 
}
\end{lstlisting}

This procedure is executed, if the sub-goal \emph{monitor\_traffic} is active.
This leads to the addition of the sub-goals
\emph{observe\_blind\_spot\_warning}, \emph{observe\_windshield} and
\emph{observe\_left\_mirror} to the active sub-goals. At this point
\emph{monitor\_traffic} is completed and the two newly added sub-goals will be
processed in the next simulation step.

One major advantage of cognitive architectures is that they provide a software
framework to simulate models of human behavior. The structure and function of
these architectures are based on findings from cognitive science. Additionally
most cognitive architectures contain several modules like the memory which
contains the above described procedures and the declarative memory for facts.
Additional modules could be included to simulate human perception and motor
control. In the above described example the perception module could also
simulate the gaze direction from the windshield to the left mirror and back and
the motor module could simulate the hand movement required for steering. Since
the time required for such actions is also based on results from cognitive
research these simulations can be used to predict the time needed for a given
task. Such cognitive driver models comprising different abstraction levels have
been investigated in \cite{liu2006driver}, \cite{salvucci2006modeling} and
\cite{ludtke2009modeling}.
Another important application of these cognitive models is the modelling of
human error. \cite{ludtke2009modeling} investigated human error in lane merging
tasks and aviation and \cite{ludtke2010cognitive} proposed a cognitive pilot
model that simulates the process of Learned Carelessness. Learned Carelessness
corresponds to ``effort-optimizing  shortcuts  leading  to  inadequate
simplifications  that  omit  safety  critical  aspects of normative
activities''.

\section{Example}
\label{sec:example}
\begin{figure}
    \centering
 	\includegraphics[height=4cm]{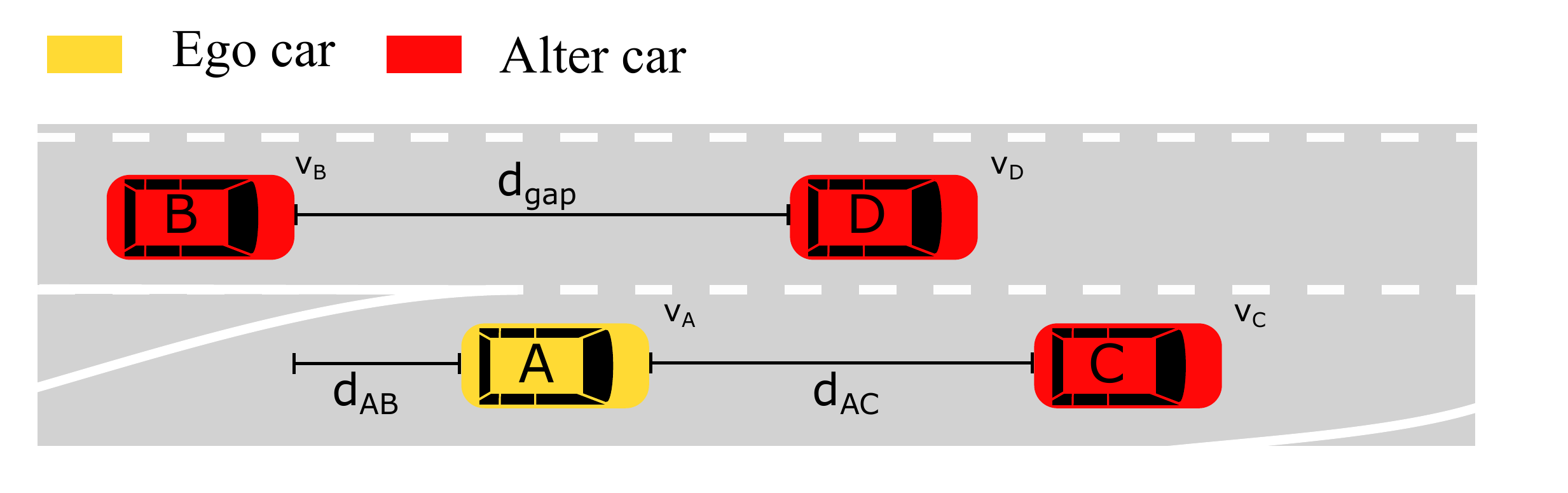}
	\caption{A simple freeway lane merging example. The ego vehicle $A$ wants 
	to merge onto the first lane of the freeway. $A$ has to keep its distance to
	the car in front of it and simultaneously looking for a	gap to merge onto 
	the freeway.}	
	\label{fig:freeway_merging}
\end{figure}
As an example, we will consider a freeway merging situation as depicted in
Figure~\ref{fig:freeway_merging}. The ego vehicle ($A$) is driving on a freeway
ramp with one alter vehicle in front it
($C$).  The distance between the ego vehicle and $C$ is called $d_{AC}$. $A$
wants to merge onto the first lane of the freeway. We consider two alter
vehicles B and D driving on this lane, whereas $B$ is the rear and $D$ the front
car. The distance between $B$ and $D$ is named $d_{gap}$ and the distance
between $A$ and $B$ is called $d_{AB}$. Every car in this situation can have a
variable velocity which leads to difference velocities like $v_{diff_{AB}} =
v_{B} – v_{A}$.  For a successful merging maneuver the driver in the ego vehicle
has to achieve several sub-goals like the selection of acceleration or check the
distance to the front car. The relevant sub-goals and their interconnection will
be described in Section~\ref{sec:driver_model}.  For this example we assume that
the ego car will be equipped with a collision avoidance system (CAS). 

In the first scenario, the ego car $A$ rear-ends the leading car $C$. For this
scenario to happen, both the CAS needs to malfunction and the human driver
needs to make a mistake. In this discussion we assume that these are the only
two possible causes; so we assume that we can produce evidence that other
causes, like a total brake failure, can be ruled out. 

In the second scenario we assume that the ego car, $A$, collides with alter car
$B$. In contrast to the scenario above, where the CAS can avoid a collision with
the lead car, we have no pure technical means to avoid a collision. Since most
highway laws require the merging car to yield, if car $A$ collides with car
$B$, it was always the fault of the driver of $A$. However, we assume that car
$A$ has a blind spot warning system, that should alert the driver to the
presence of car $B$.

\subsection{Technical Models}
\begin{figure}%
    \centering
    \begin{subfigure}{7cm}
        \includegraphics[height=9cm]{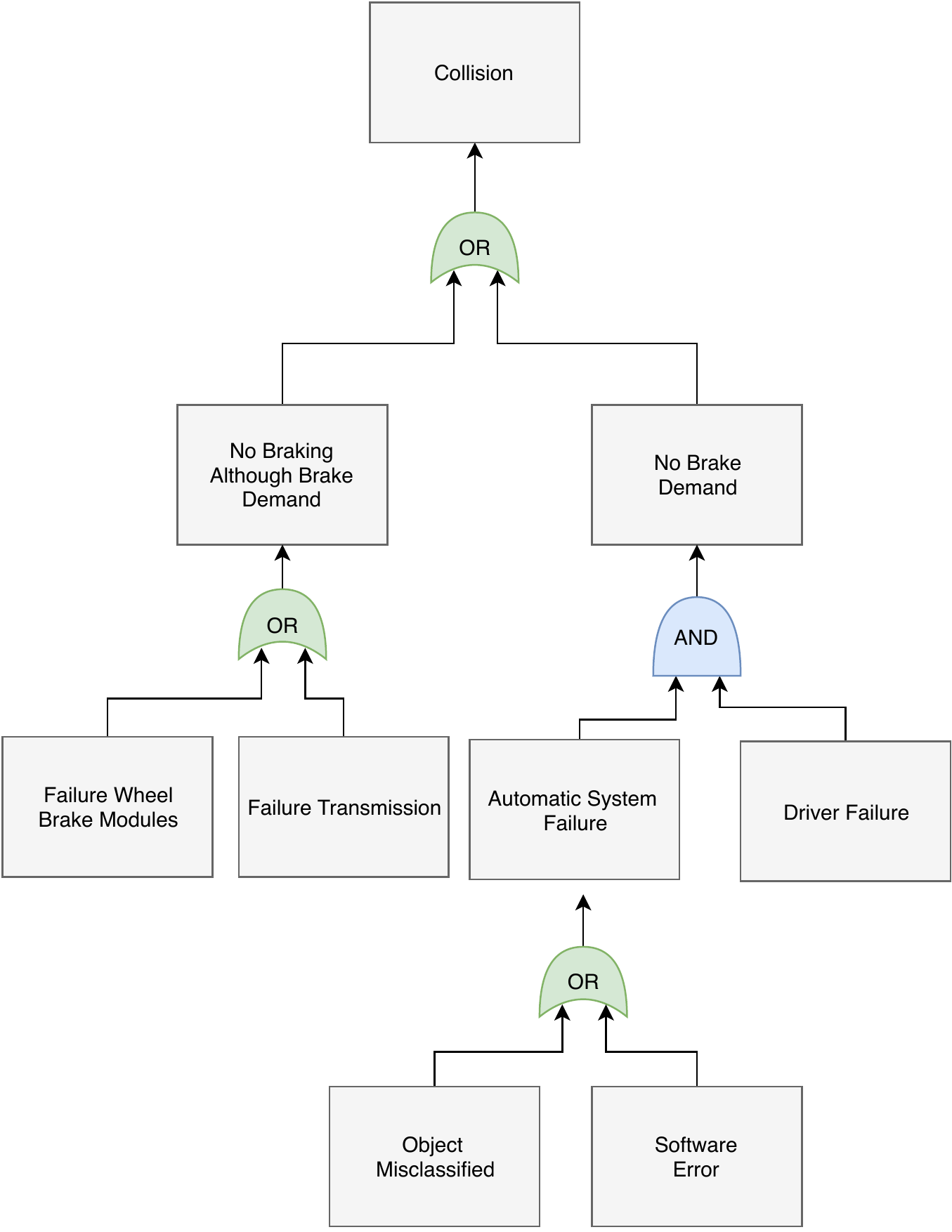}
        \caption{Technical.}	\label{fig:fault_tree}
    \end{subfigure}
    \qquad
    \begin{subfigure}{5cm}
        \includegraphics[height=9cm]{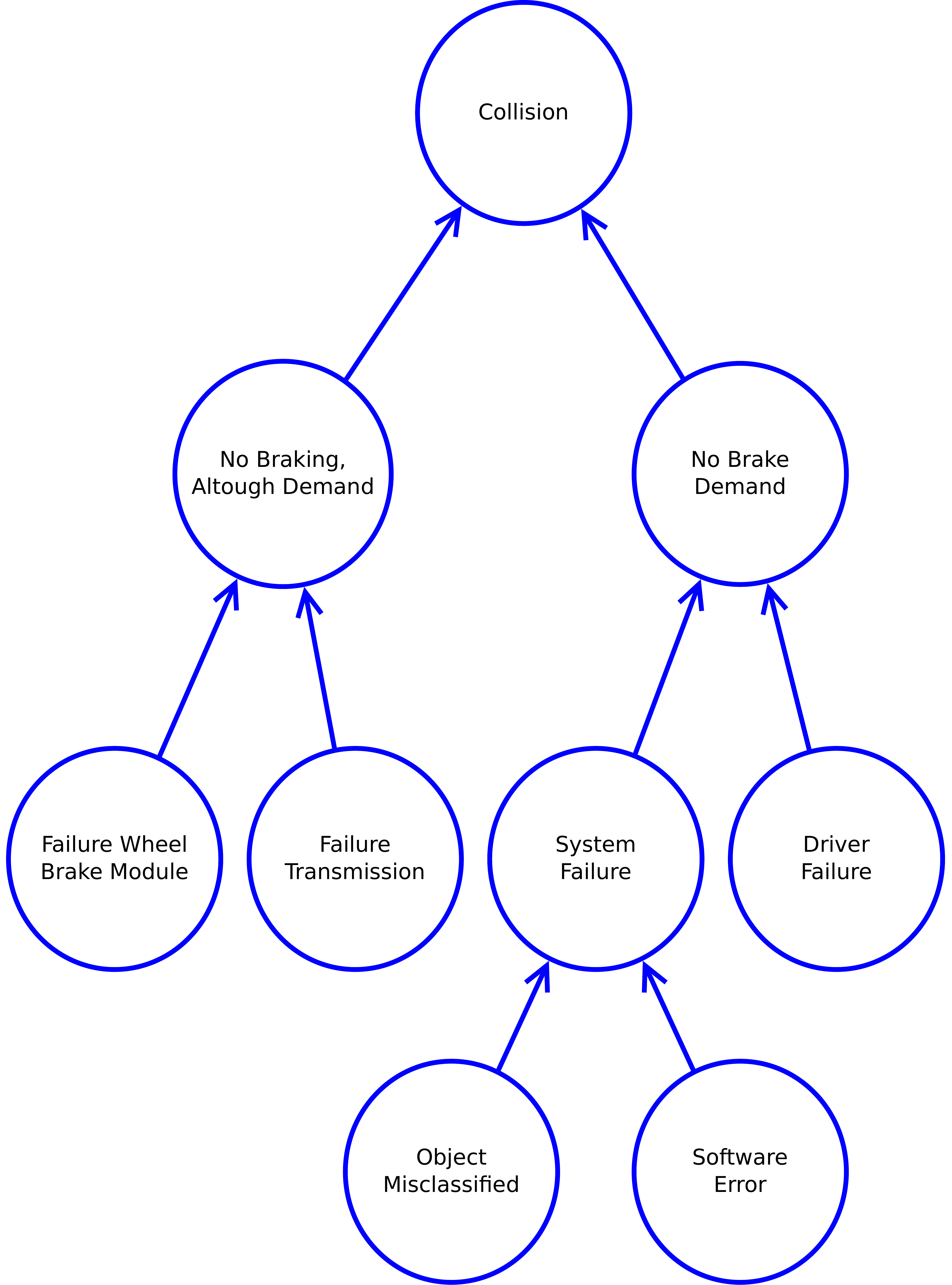}
        \caption{Causal.}
		\label{fig:fault_tree_causal}
    \end{subfigure}
    \caption{Fault tree.}%
\end{figure}

\begin{figure}%
    \centering
    \begin{subfigure}{5cm}
        \includegraphics[height=9cm]{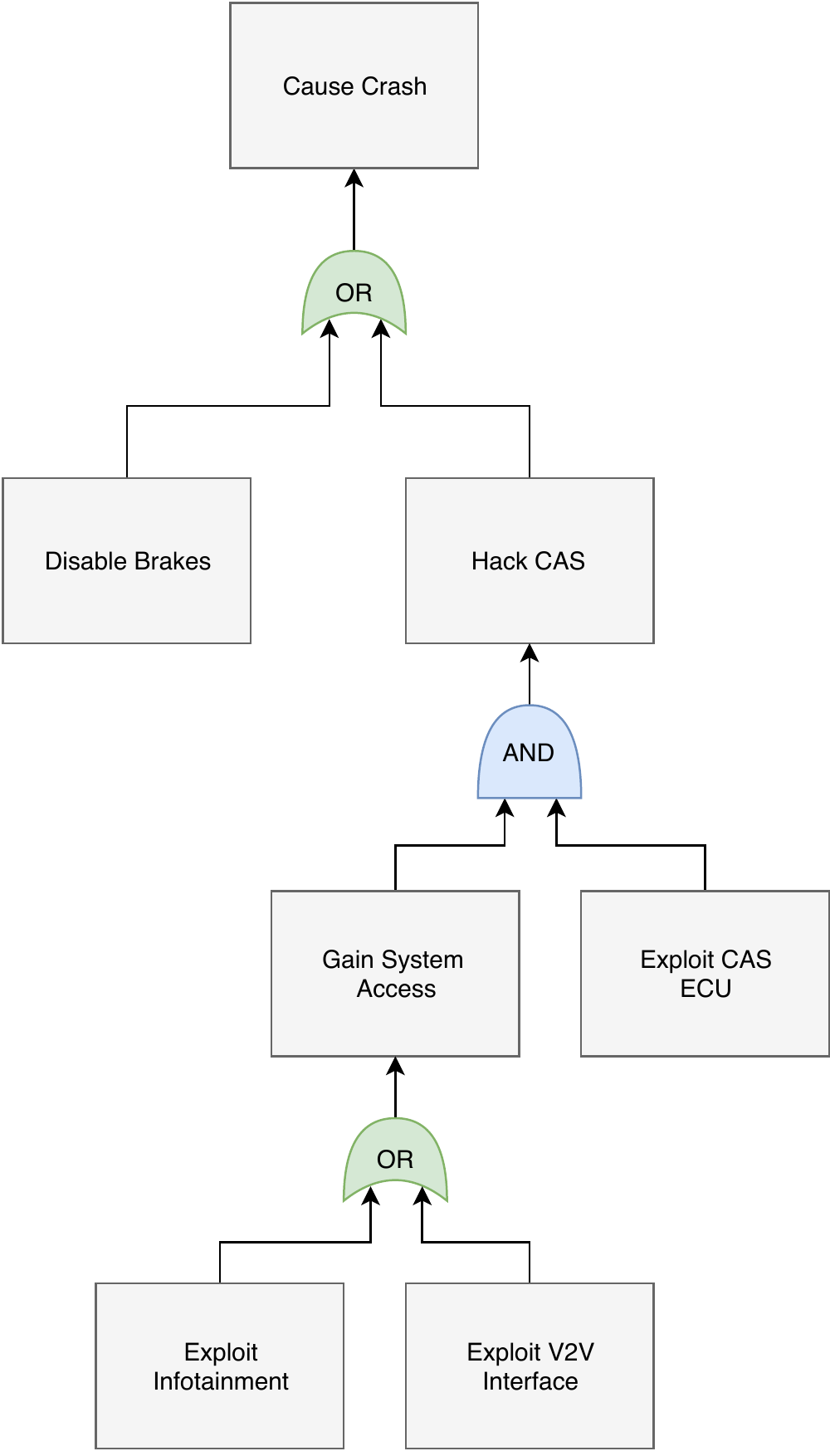}
        \caption{Technical.}
        \label{fig:attack_tree}
    \end{subfigure}
    \qquad
    \begin{subfigure}{5cm}
        \includegraphics[height=9cm]{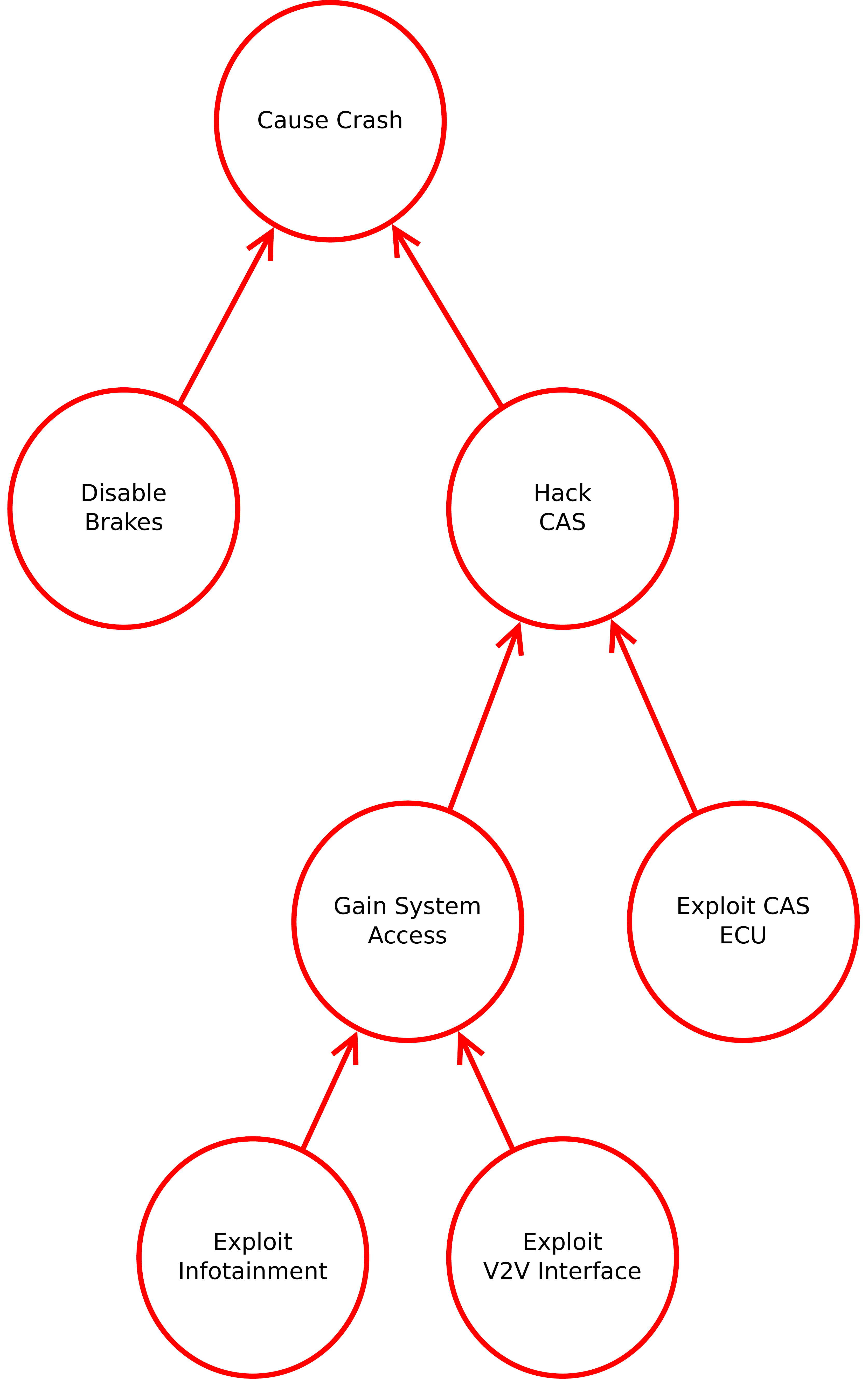}
        \caption{Causal.}
        \label{fig:attack_tree_causal}
    \end{subfigure}
    \caption{Attack tree.}%
\end{figure}
A major problem for causal reasoning is the lack of models. Creating causal
models manually is a tedious process and impossible without detailed technical
knowledge of the system that is being modeled. To kickstart this process, we
have shown in \cite{nfm} that it is possible to use existing models of systems
to seed a causal model. Having an automatically generated model as a base or
scaffold for causal models makes it easier for domain experts to build the
models. Additionally, resulting causal models can then be reused in future
investigations that analyze similar accidents. In \cite{nfm} we have shown how
attack trees, fault trees and timed failure propagation models can be converted
to a causal model and be merged into a holistic causal model. To keep the
paper focused, we will restrict the discussion to attack and fault trees.
However, in practice many different system models could be used to seed the
causal models. 

Figure~\ref{fig:fault_tree} shows the fault tree for our example. It is based on
a fault tree provided by \cite{isermann2002fault} and depicts reasons why a car
might not brake and collide with another car. One branch, \emph{No Braking
Although Brake Demand} covers mechanical failures of the brake system, while the
other branch, \emph{No Brake Demand}, covers cases in which the brakes are
working, but no brake attempt is being made. 

Figure~\ref{fig:attack_tree} likewise present a simplified attack tree for our
example. The left branch, \emph{Disable Brakes}, covers cases where an attacker
manages to disable the brakes. Here we do not detail possible ways to do so any
further, but they can reach from software attack to the classic ``cut the brake
lines''. The right branch covers some ways in which an attacker could attack the
CAS subsystem and cause a crash. This attack tree is modeled on the widely
publicized ``Jeep Hack'' \cite{jeep}.  

\subsection{Driver Model}
\label{sec:driver_model}

\begin{figure}%
    \centering

    \begin{subfigure}{12cm}
	 	\includegraphics[height=7cm]{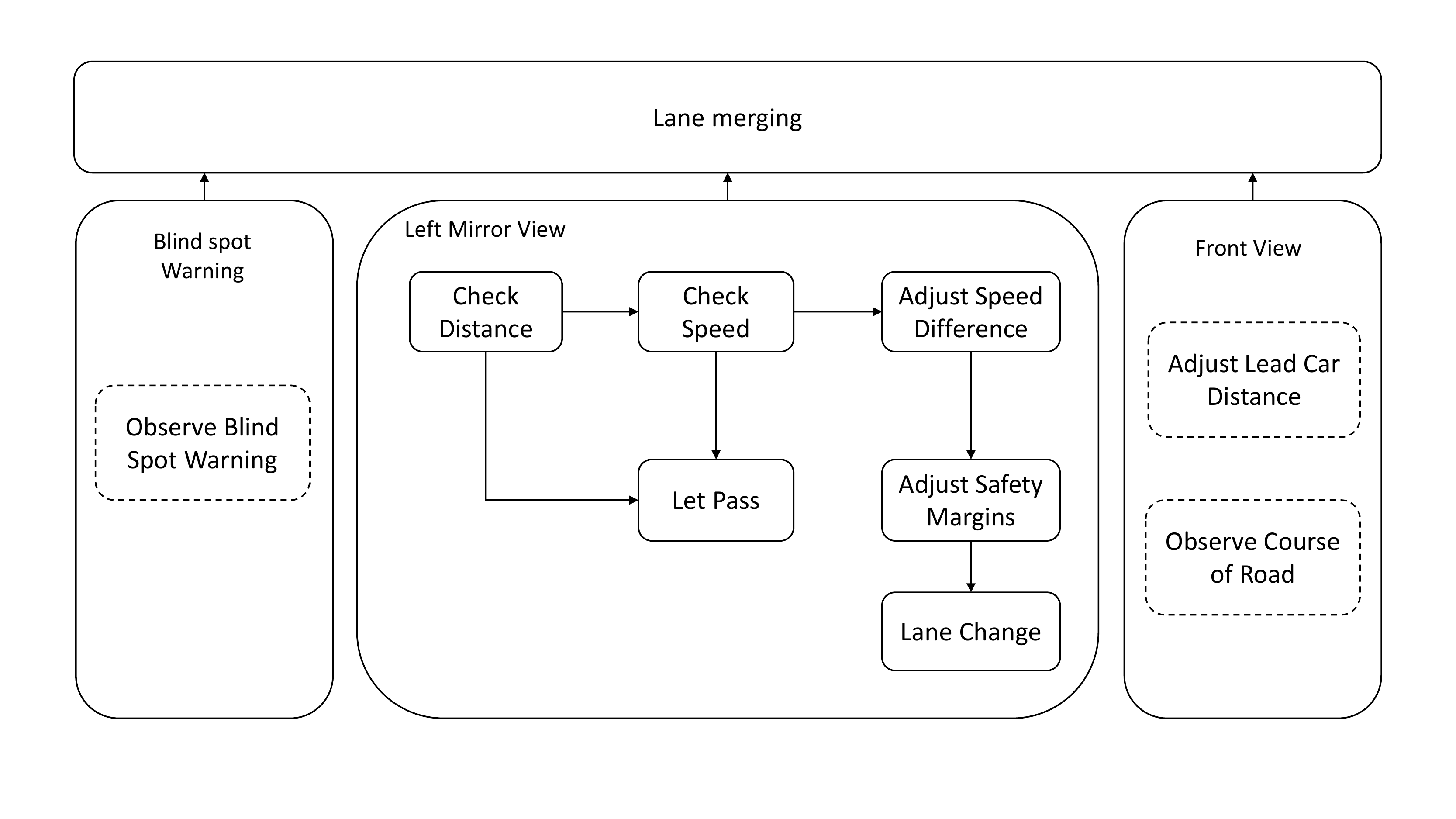}
		\caption{HTA model.}	
		\label{fig:state_chart}
    \end{subfigure}

    \begin{subfigure}{12cm}
        \includegraphics[height=9cm]{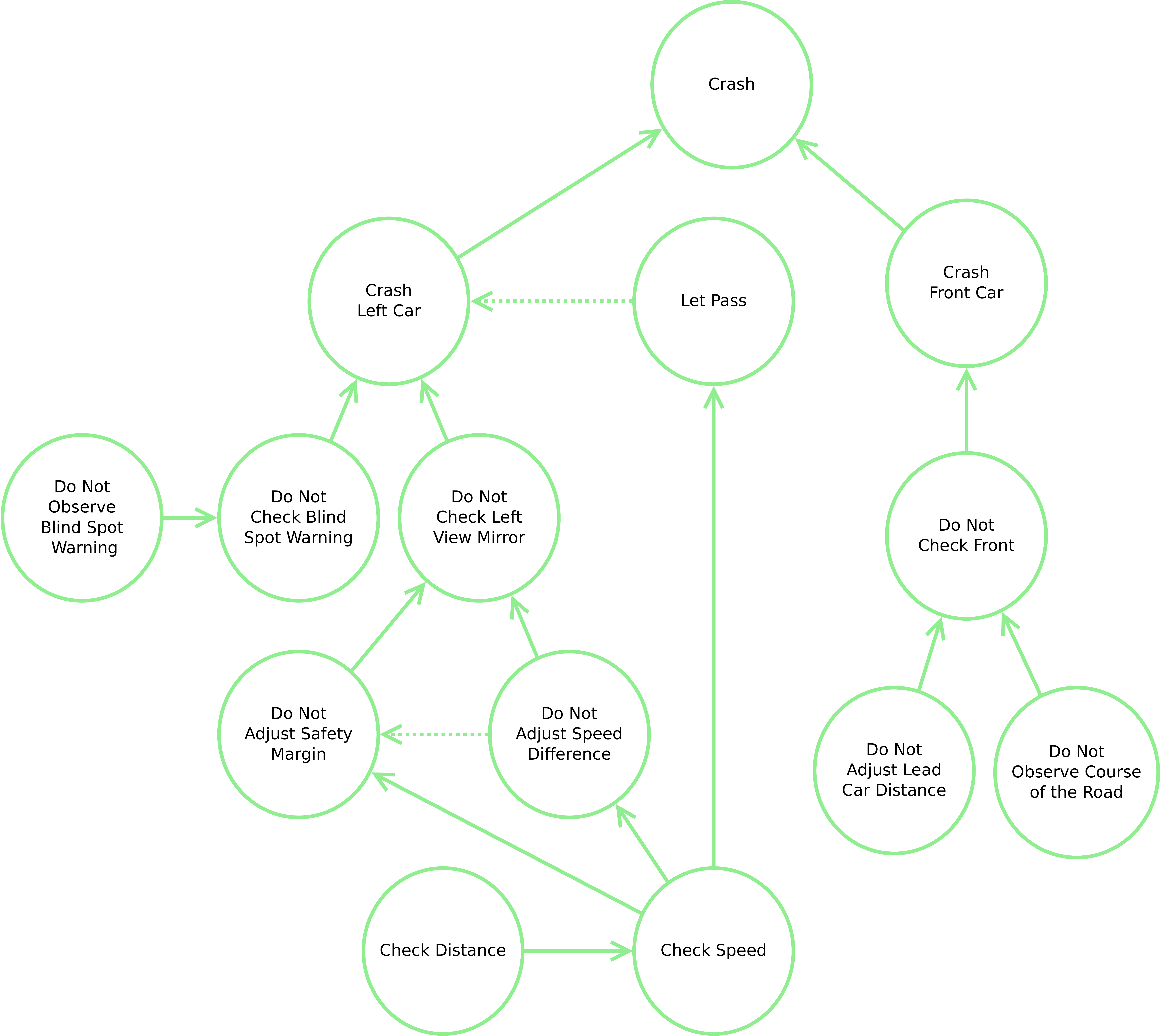}
        \caption{Causal Model.}
		\label{fig:hta_causal}
    \end{subfigure}
    \caption{Modified Driver Model based on \cite{ kassner2011hierarchical}.}%
\end{figure}

We use the structure of the driver model for freeway lane change tasks presented
in \cite{ weber2009modellierung} and \cite{kassner2011hierarchical} as a basis
for our human driver model. A slightly modified version of the state chart from
\cite{ kassner2011hierarchical} can be found in Figure~\ref{fig:state_chart}.
Here the parent goal of the Lane Change Manoeuvre Task has three main sub-goals:
observing the blind spot warning, observing the left mirror view and observing
the front view through the windshield.

The blind spot warning sub-goal has one further sub-goal. During a lane merging
maneuver the driver has to be sure that no car is driving in the blind spot. In
this example we assume that the driver only uses the installed blind spot
warning system for this goal.  The front view sub-goal consists of two
sub-goals, namely adjusting the distance to the lead car $C$ and observing the
course of the road.  The second sub-goal describes the observation of the alter
vehicles through the left mirror. The first sub-goal of the driver model is the
estimation of the distance between the car $B$ on the freeway and the ego
vehicle $A$. If the distance $d_{AB}$ is big enough the driver will check the
speed $v_{B}$ and estimate, based on a threshold if the relative velocity
$v_{AB}$ is low enough for a merging maneuver. If the conditions for one of
these two goals are not fulfilled the driver will let $B$ pass and start with
the first goal again. If $v_{AB}$ is lower than the threshold, the driver’s
confidence that the lane change will be successful rises and $Adjust Speed
Difference$ will be triggered. This means that the driver will accelerate or
decelerate to have the optimal relative speed w.r.t.  $B$ and $D$ for the lane
change. The last sub-goal is the adjustment of the safety margin to all adjacent
cars during the lane changing. This last sub-goal completes the lane change
goal. 

\subsection{Setting the Context}

When converting technical models, like fault- or attack trees, or human models,
like HTA models, to causal models we need to be mindful of the fact that these
models describe \emph{type causality}, while we try to reason about \emph{actual
causality}. Following Halpern \cite{halpern2016actual}, type causality is
``typically forward-looking and used for prediction'', whereas actual causality
``retrospectively ask[s] why [something] occurred''. In our example, fault
trees, attack trees, and HTA models are created from measurements, predictions
and expert judgment. To reason about actual causality, however, we need to link
those models to actual events and actors. In general, this is a lot easier for
technical systems and often next to impossible for human beings. The reason is
that we can instrumentalize and monitor technical systems with arbitrary
resolution and precision, whereas observations of humans are incomplete and
fuzzy. For example, we might have eye-tracking sensors that allows us to deduce
that a driver was about to merge into the left lane; if we analyze the same
scenario on a motorcycle, on the other hand, we most likely cannot see the
rider's eye under the helmet and are unable to deduce any
driver intentions at all. In a technical model, in contrast, we could in both
instances detect if the turn indicator light was set. Both, the technical and
the human model, are similar in structure, but very  different in the available
level of logging granularity.  Highly automated vehicles (HAVs) are uniquely suitable to
investigate the integration of human and technical models, because the gap
between the log granularity is comparatively small. HAVs have access to a wide
array of very precise sensors, are equipped with the computing power to process
the data, have the storage capacity to record them, and, most importantly, have
driver monitoring systems to gather reliable ``log data'' about the driver.
An additional advantage in the HAV example is that the social context is very
well defined. All drivers go to driving school and learn similar patterns of
behavior there. In common situations the drivers will react highly
similarly. Defining the context for a less standardized task, e.g., writing a 
poem, is a hard challenge for future research.

\section{Transforming and Joining the Models}
\subsection{Transforming Trees into Causal Models}\label{subsec:step1}

Comparing Definition~\ref{def:aft} and Definition~\ref{def:cm}, the mapping is
straightforward. Again following \cite{nfm}, Definition~\ref{def:aft_cm} shows
that we transform each node in a tree into an \textit{endogenous variable} that
determines whether or not a specific event occurred. For now, we only
consider binary causal models and express the operators of the tree in the  
structural equations.

To fulfill the requirement that each endogenous variable is defined by the other
endogenous or exogenous variables \cite{halpern2005causes1},  we define an
exogenous variable (has the same name as the tree node with an ``\_exo'' suffix)
that will provide the value of the node's endogenous variable. This turns those
leafs into endogenous and allows us to identify leaf nodes as causes of events.

\begin{definition}\label{def:aft_cm}
	\textbf{Attack/Fault Tree To Causal Model}\\
	 $T = \mathcal{(N,\to, }n_0)$ is mapped to a  $M = \mathcal{((U,V,R),F)}$, 
	 i.e., $T \rightarrow M$ as follows 
	\begin{itemize}
		\item $\mathcal{U}= E(T,\_exo)$, where $E(T,suffix)$ returns a renamed
		copy of the leaf nodes of a tree $T$ with a suffix ``\_exo''.
		\item $\mathcal{V}= N$. 
		\item $\mathcal{R}= \{0,1\}$.
		\item $\mathcal{F}$ associates with each $X \in V-E(T)$ a propositional 
		formula based on the tree gates; and with each $X \in E(T)$ a formula of 
		the form $X=X$\_exo. 
	\end{itemize}
\end{definition}

\begin{figure}
    \centering
        \includegraphics[height=16cm]{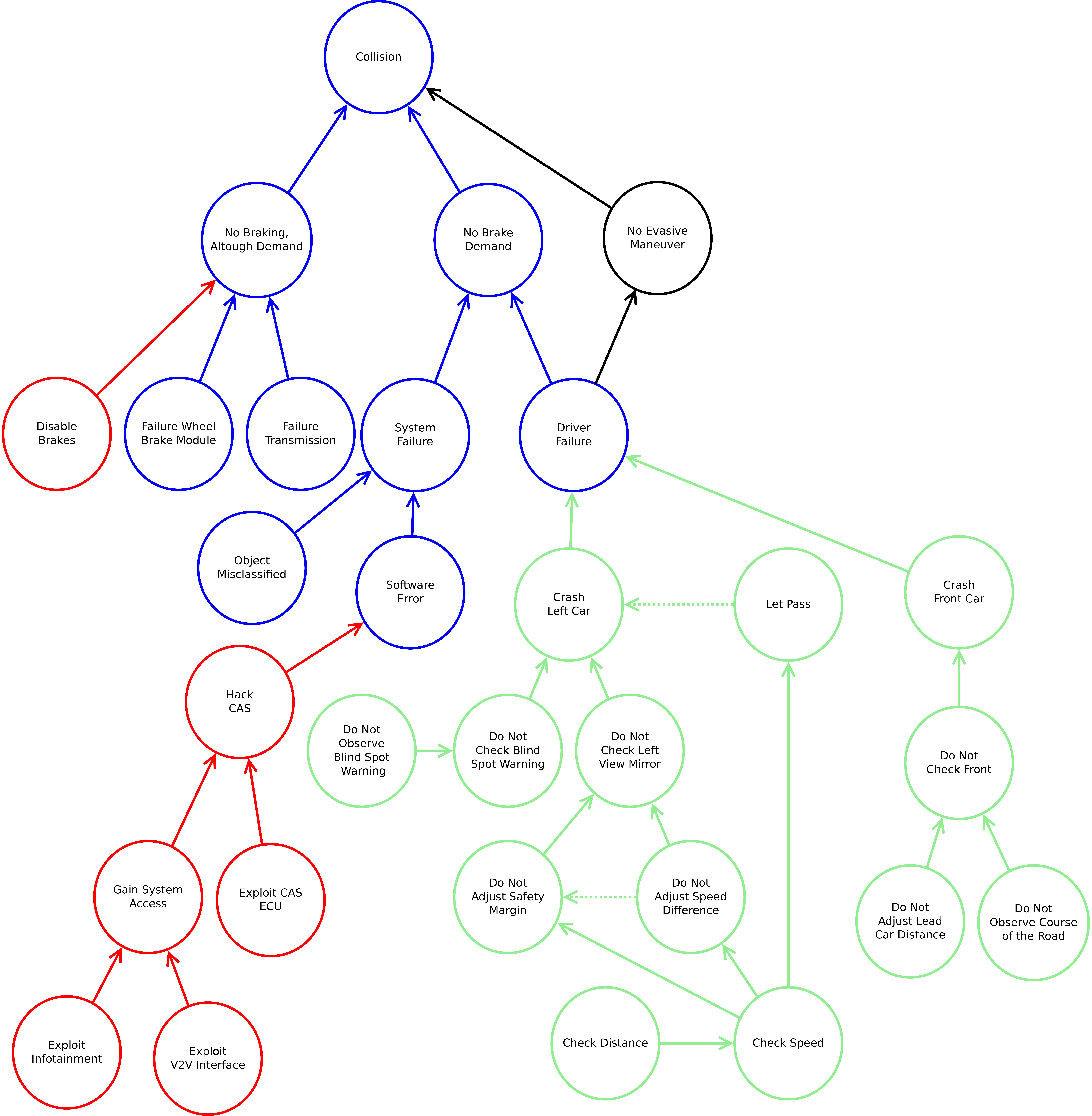}
        \caption{Integrated socio-technical causal model.}
        \label{fig:combined_causal}
\end{figure}

\subsection{Transforming Human Models into Causal Models}
The presented model of human behavior differs from the technical models in one
important way: It models positive behavior. The technical models in this paper,
as well as in previous work \cite{nfm}, describe things that can go wrong. The
presented human models, in contrast, describes the typical positive behavior. So
if we want to analyze a crash, or other negative events, we first need to
transform the positive model into a negative model. In
Figure~\ref{fig:state_chart}, for example, a crash can happen if the driver does
not check the blind spot warning and does not check the left mirror and does not
check the front view.  

Due to the intricacies of inverting only some nodes, we created the causal model
in Figure~\ref{fig:hta_causal} manually. To convert the HTA model to a causal
model, we first started by modeling that the HTA actually shows actions to
prevent two different crashes. The blind spot warning and the left view mirror
prevent crashes with cars in the first highway lane and checking the front view
prevents against read-ending leading cars on the ramp. The right hand sub-tree,
that prevents rear-ending a lead car, is fairly easy to model. If the driver
does not check the front view, the car will crash into any lead vehicle. This
check is done by two sub-goals: adjusting the distance to any lead car and
observing the stretch of road ahead. 

Avoiding a crash with a car in the first lane of the highway can be caused by
omitting two actions: checking the blind spot warning and not checking the left
view mirror. When checking the left view mirror, adjusting the speed will
preempt (dotted lined arrow) adjusting the safety marking. Additionally, a crash
with a car in the first lane of the highway can be preempted by simply letting
it pass.  

\begin{figure}
\small
\begin{framed}
$Collision = NoBrakingAltoughDemand \lor NoBrakeDemand \lor NoEvasiveManeuver$\\
$ NoBrakingAltoughDemand = DisableBrakes \lor FailureWheelBrakeModule
\lor FailureTransmission$\\
$NoBrakeDemand = SystemFailure \lor DriverFailure$\\
$NoEvasiveManeuver=DriverFailure$\\
$DriverFailure =  CrashLeftCar \lor CrashFrontCar$\\
$SystemFailure = ObjectMissclassified \lor SoftwareError$\\
$CrashLeftCar = (DoNotCheckBlindSpotWarning \land DoNotCheckLeftViewMirror)
\land $  \dbox{$\neg LetPass$}\\
$LetPass = CheckSpeed$\\
$CrashFrontCar = DoNotCheckFront$\\
$DoNotCheckBlindSpotWarning = DoNotObserveBlindSpot$\\
$DoNotCheckLeftViewMirror = DoNotAdjustSafetyMargin \land
DoNotAdjustSpeedDifference $\\
$DoNotAdjustSafetyMargin = CheckSpeed \land $ \dbox{$\neg
DoNotAdjustSpeedDifference$}\\
$DoNotAdjustSpeedDifference = CheckSpeed$\\
$CheckSpeed = CheckDistance$\\
$DoNotCheckFront = DoNotAdustLeadCarDistance \land
DoNotObserveCourseOfTheRoad$\\
$SoftwareError = HackCAS$\\
$HackCAS = GainSystemAccess \land ExploitCASECU$\\
$GainSystemAccess = ExploitInfotainment \lor ExploitV2VInterface$
\end{framed}
\caption{The equations for the integrated model. Dashed Boxes highlight 
preemption relations. }
\label{fig:equations}
\end{figure}

\subsection{The Integrated Socio-Technical Model}

Figure~\ref{fig:combined_causal} depicts the combined causal model and
Figure~\ref{fig:equations} the corresponding structural equations. The source
models were joined similar to the procedure described in \cite{nfm} and can now
be used to reason about the actual causality in concrete accidents or events. In
this concrete example, we started with the fault tree and then first joined it
with the attack tree. In the process, we split the attack tree and connected
\emph{Disable Brakes} with the fault \emph{No Braking Although Demand}.
\emph{Hack CAS} was then connected to the \emph{Software Error} fault. To join
the HTA model, we equated its top node, \emph{Crash}, with the fault
trees \emph{Driver Failure}. The reason for this choice is that when viewed
under the prospective of a crash, all human actions are seen as either correct,
or as wrong and being a human error. 

To set the context (i.e., the value of each variable), we of course need sensors
and monitors to provide us with the values. The integrated model is useful
in two ways: it helps us to decide what sensors and log data we need and it
allows us to reason over the results. The first part is unfortunately less
useful than might be at first expected. Since the causal model originates from
technical models, it only considers states for which there anyway are
sensors or at least expectations of problems. The same holds true for the human
model: it generally only considers what can be measured in a lab. It is slightly
more insightful, because it can inform developers of HAVs which driver sensors
are relevant and which data is necessary. 

The real benefit of the integrated socio-technical causal model is that it can
be used by experts to guide their investigation. One advantage is that they can
extend the model with knowledge from other sources. The model in
Figure~\ref{fig:combined_causal}, for example, was extended with the node
\emph{No evasive maneuver} (colored in black). The core advantage, however, is
that we can reason across multiple models. For example, we can evaluate
scenarios in which the driver did not check the left view mirror and at the same
time the CAS did not detect a car in the highway lane. We can also express that
in scenarios, like the Jeep hack, where an attacker has the ability to disable
the brakes, both the CAS and the human driver could not have prevented an
accident.  In our experience, causal model are uniquely suitable for such cross
domain modeling and reasoning.

\begin{figure}%
    \centering

    \begin{subfigure}{5cm}
	 	\includegraphics[height=6.75cm]{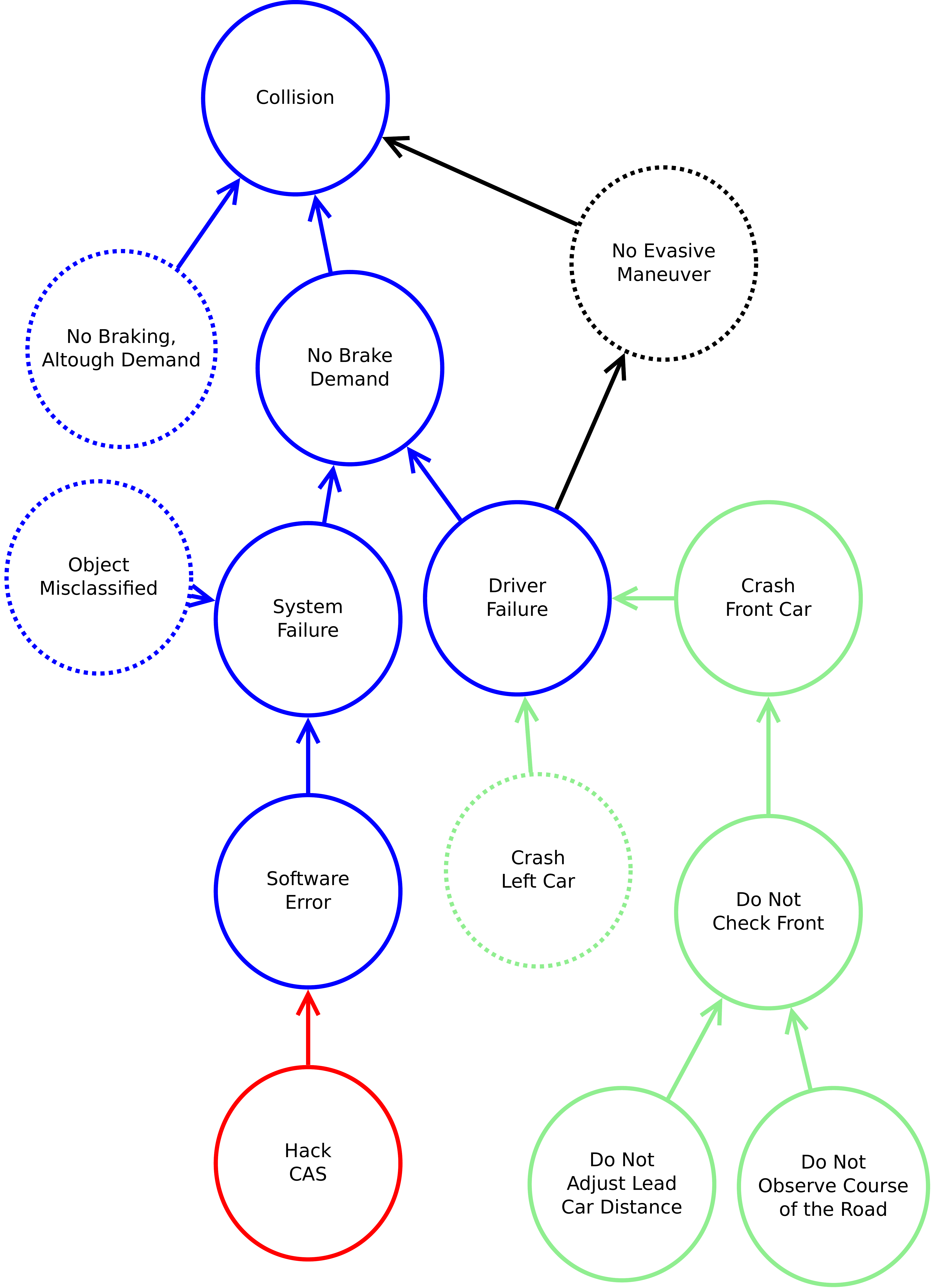}
		\caption{Scenario 1.}	
		\label{fig:scenario1}
    \end{subfigure}
	\qquad
    \begin{subfigure}{5cm}
        \includegraphics[height=6.75cm]{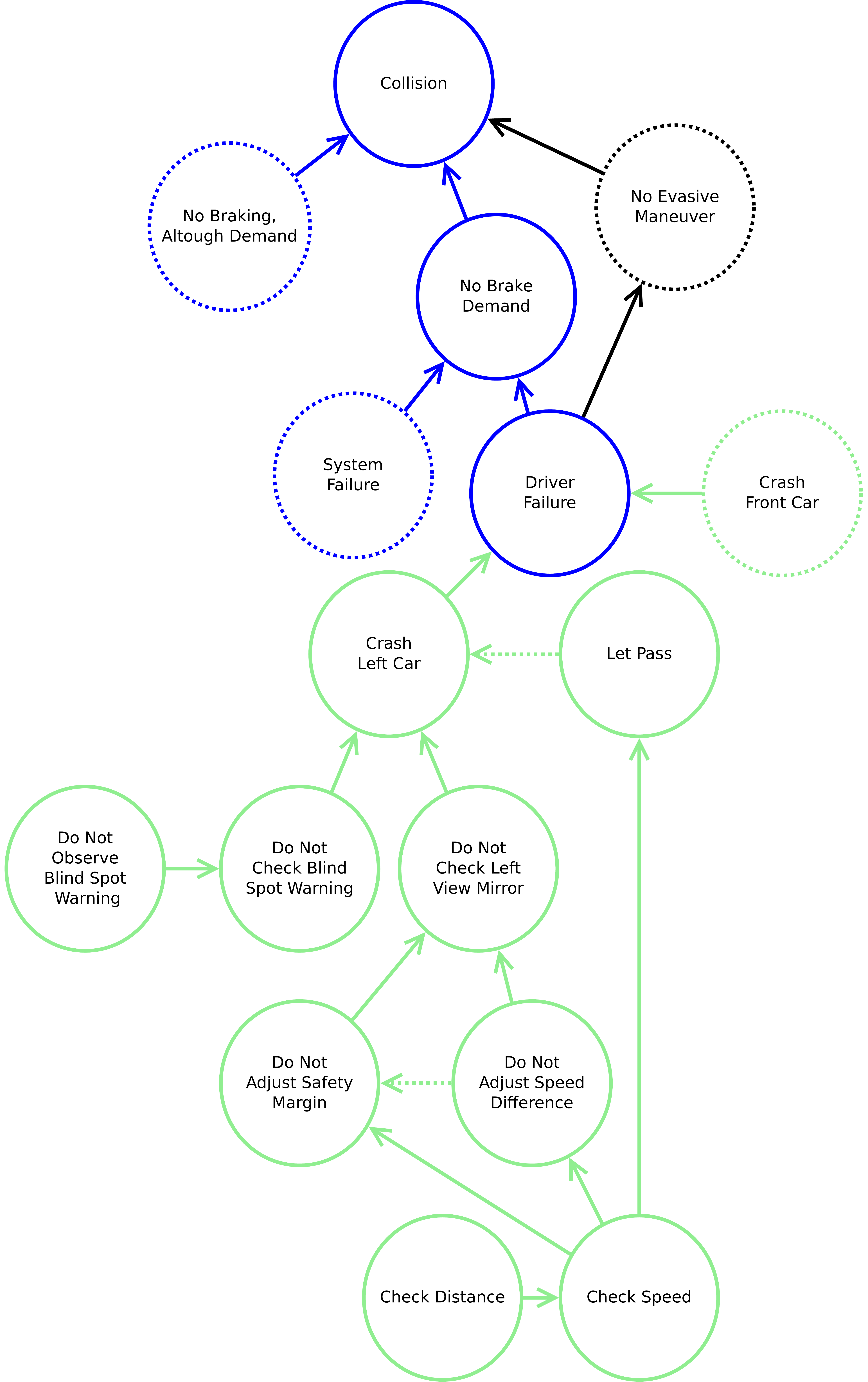}
        \caption{Scenario 2.}
		\label{fig:scenario2}
    \end{subfigure}
    \caption{Simplified causal models.}%
\end{figure}

\subsection{Analyzing Scenarios}
We can now use the integrated socio-technical causal model
(Figure~\ref{fig:combined_causal}) to reason about the two scenarios of our
example (Section~\ref{sec:example}). If there was a rear-end collision and we
can confirm that there was a demand to brake, we can, in a first step,
disregard the left most sub-tree under the node \emph{No Braking, Although
Demand}. Since we crashed into the lead car, we can also ignore the sub-tree
\emph{Crash Left Car} from the HTA model. If we can now find evidence for a
hacking attempt against the CAS, we know that a combination of driver error and
software error, inducted by the hacking is the cause of the event.
Figure~\ref{fig:scenario1} depicts this causal model, with showing ``pruned
nodes'' with dashed borders.  We can now also model a preemption relation
between the \emph{system failure} and the \emph{driver failure}. By choosing the
direction of this preemption, we can model our view of autonomy: if the driver
preempts the system, we are talking about non autonomous cars, in which the
driver needs to always pay attention and if the system preempts the driver, we
model fully autonomous vehicles that do not require driver attention. There is,
of course, no right or wrong model -- the correctness depends on the context.
The great advantage of causal model is that they can capture those assumption
and explicate them for further discussion. 

Figure~\ref{fig:scenario2} depicts the causal model for the scenario in which
the car crashes into the car on the first lane of the highway. In this scenario
we have no active technological measure to avoid the crash -- the fault arises
from a human error. From the model, we can deduce that the driver did not look
(or ignored) the blind spot warning light and did not check the left mirror.
Also the driver did not let the other car pass, thus not preempting the crash.

\section{Conclusion}
In this paper, we have shown how we can use Halpern-Pearl causal models as a
\emph{lingua franca} to combine technical and social models of a single system
into an integrated socio-technical causal model. The expressive power and
versatility of causal models allows us to transform the source models and, in
some cases even automatically, combine them into a single model. The major
advantage of this approach is that we can reuse already existing models to seed
the causal models and do not have to build them from scratch. The integrated
model can then be revised by experts and used to reason over problems that cross
multiple models. 

Our work, however, is still in its infancy and poses several challenges for
future work. While the joining of models can be done automatically in many
cases, there are many instances where we still need expert intervention;
especially joining positive and negative models automatically is a great
challenge. Converting the human models into causal models is still a wide open
problem.  For one, developing those human models requires time consuming and
expensive empirical studies. Since negative situations, such as crashes, are,
even in simulations, relatively rare, building and validating negative models of
human behavior is significantly harder than developing positive models. Another
major issue is how to actually set the context for human models, or, in other
words, how can we measure the human behavior in sufficient detail. While
generalized, type causal, knowledge is useful in research, for actual causality
we need to know exactly what a certain individual has done. While recording this
data is a formidable technical challenge, it also raises very important issues
about privacy and data ownership. One solution could be to rely on cognitive
architectures, connected to detailed HAV simulators to analyze actual and
possible accident scenarios. 

Despite these open challenges, we are convinced that extending technical models
with human models and use automatic tool to reason about these models is an
essential building block for accountable CPS that integrate well into their
societal context. The example given in this paper highlights the future
potential and the general feasibility of our approach.\\  
\noindent
\textbf{Acknowledgments.} This work was supported by the  Deutsche
Forschungsgemeinschaft (DFG) under grant no. PR1266/3-1, Design Paradigms for
Societal-Scale Cyber-Physical Systems. 
\bibliographystyle{eptcs}
\bibliography{bibliography}
\end{document}